\documentclass[letterpaper, 10 pt, conference]{ieeeconf}  % Comment this line out if you need a4paper

\IEEEoverridecommandlockouts                              % This command is only needed if 
\usepackage{graphicx} % for handling graphics
\usepackage{amsmath, amssymb, amsfonts} % for math
\usepackage{mathtools, commath} % for additional math functionality
\usepackage{url} % for URLs
\usepackage{float} % for controlling float placement
\usepackage[noend]{algpseudocode} % for pseudocode
\usepackage[ruled,vlined]{algorithm2e} % for algorithms (alternative to algpseudocode)
\RestyleAlgo{ruled} % style for algorithm2e
\usepackage{hyperref} % for clickable links
\usepackage[table]{xcolor} % for table colors
\usepackage{multirow} % for multi-row cells in tables
\usepackage{subcaption} % for subfigures and subtables
\usepackage{balance} % for balancing references
\usepackage{tikz}
\usepackage{MnSymbol}

\title{\LARGE \bf
Model Predictive Path Integral Control of I$^2$RIS Robot Using RBF Identifier and Extended Kalman Filter*
}

\author{Mojtaba Esfandiari$^{1}$, Pengyuan Du$^{1}$, Haochen Wei$^{2}$, {\it Graduate Student Member, IEEE},  Peter Gehlbach$^{4}$, \\ Adnan Munawar$^{3}$, Peter Kazanzides$^{2}$, {\it Member, IEEE}, Iulian Iordachita$^{1}$, {\it Senior Member, IEEE}% <-this % stops a space
\thanks{*This work was supported by the U.S. National Institutes of Health under grant numbers R01EB023943 and R01EB034397 and partially by JHU internal funds.}% <-this % stops a space
\thanks{$^{1}$ Mojtaba Esfandiari, Pengyuan Du, and Iulian Iordachita are with the Department of Mechanical Engineering and Laboratory for Computational Sensing and Robotics, Johns Hopkins University,
Baltimore, MD, 21218, USA. 
        ({\tt\small mesfand2, pdu5, iordachita@jhu.edu})}%
\thanks{$^{2}$ Haochen Wei, Peter Kazanzides, are with the Department of Computer Science and Laboratory for Computational Sensing and Robotics, Johns Hopkins University,
Baltimore, MD, 21218, USA. 
        ({\tt\small hwei15, pkaz@jhu.edu})}%
\thanks{$^{3}$ Adnan Munawar is with the Laboratory for Computational Sensing and Robotics, Johns Hopkins University,
Baltimore, MD, 21218, USA. 
        ({\tt\small amunawa2@jhu.edu})}%
\thanks{$^{4}$ Peter Gehlbach is with the Wilmer Eye Institute, Johns Hopkins Hospital, Baltimore, MD, 21287, USA. ({\tt\small pgelbach@jhmi.edu})
}%
}

\begin{document}

\maketitle
\thispagestyle{empty}
\pagestyle{empty}

%%%%%%%%%%%%%%%%%%%%%%%%%%%%%%%%%%%%%%%%%%%%%%%%%%%%%%%%%%%%%%%%%%%%%%%%%%%%%%%%
\begin{abstract}

Modeling and controlling cable-driven snake robots is a challenging problem due to nonlinear mechanical properties such as hysteresis, variable stiffness, and unknown friction between the actuation cables and the robot body. This challenge is more significant for snake robots in ophthalmic surgery applications, such as the Improved Integrated Robotic Intraocular Snake (I$^2$RIS), given its small size and lack of embedded sensory feedback. 
Data-driven models take advantage of global function approximations, reducing complicated analytical models' challenge and computational costs. However, their performance might deteriorate in case of new data unseen in the training phase. Therefore, adding an adaptation mechanism might improve these models' performance during snake robots' interactions with unknown environments.   
In this work, we applied a model predictive path integral (MPPI) controller on a data-driven model of the I$^2$RIS based on the Gaussian mixture model (GMM) and Gaussian mixture regression (GMR). To analyze the performance of the MPPI in unseen robot-tissue interaction situations, unknown external disturbances and environmental loads are simulated and added to the GMM-GMR model. These uncertainties of the robot model are then identified online using a radial basis function (RBF) whose weights are updated using an extended Kalman filter (EKF). Simulation results demonstrated the robustness of the optimal control solutions of the MPPI algorithm and its computational superiority over a conventional model predictive control (MPC) algorithm.

\end{abstract}

%%%%%%%%%%%%%%%%%%%%%%%%% INTRODUCTION %%%%%%%%%%%%%%%%%%%%%
\section{INTRODUCTION}

Epiretinal Membrane (ERM) can develop due to various causes, including diabetic retinopathy, retinal vein occlusion, and ocular inflammation following vitreous separation \cite{fung2021epiretinal}. The progression of ERM leads to metamorphopsia, reduced visual acuity, and central vision loss, significantly interfering with and decreasing the quality of everyday life \cite{kanukollu2023epiretinal}. Surgeons typically perform the ERM peeling procedure manually using an inner limiting membrane (ILM) micro-forceps, needle picks, and micro-vitreoretinal blades. Multiple attempts at engaging and peeling the membrane are often necessary, which increases the risk of iatrogenic damage. Focal retinal hemorrhages are also reported due to unintentional pinches to the nerve fiber layer caused by hand tremors or human errors \cite{lumi2022simple}, and significant injuries are possible. 

\begin{figure}[t!]
    \centering
    \includegraphics[width=0.47\textwidth]{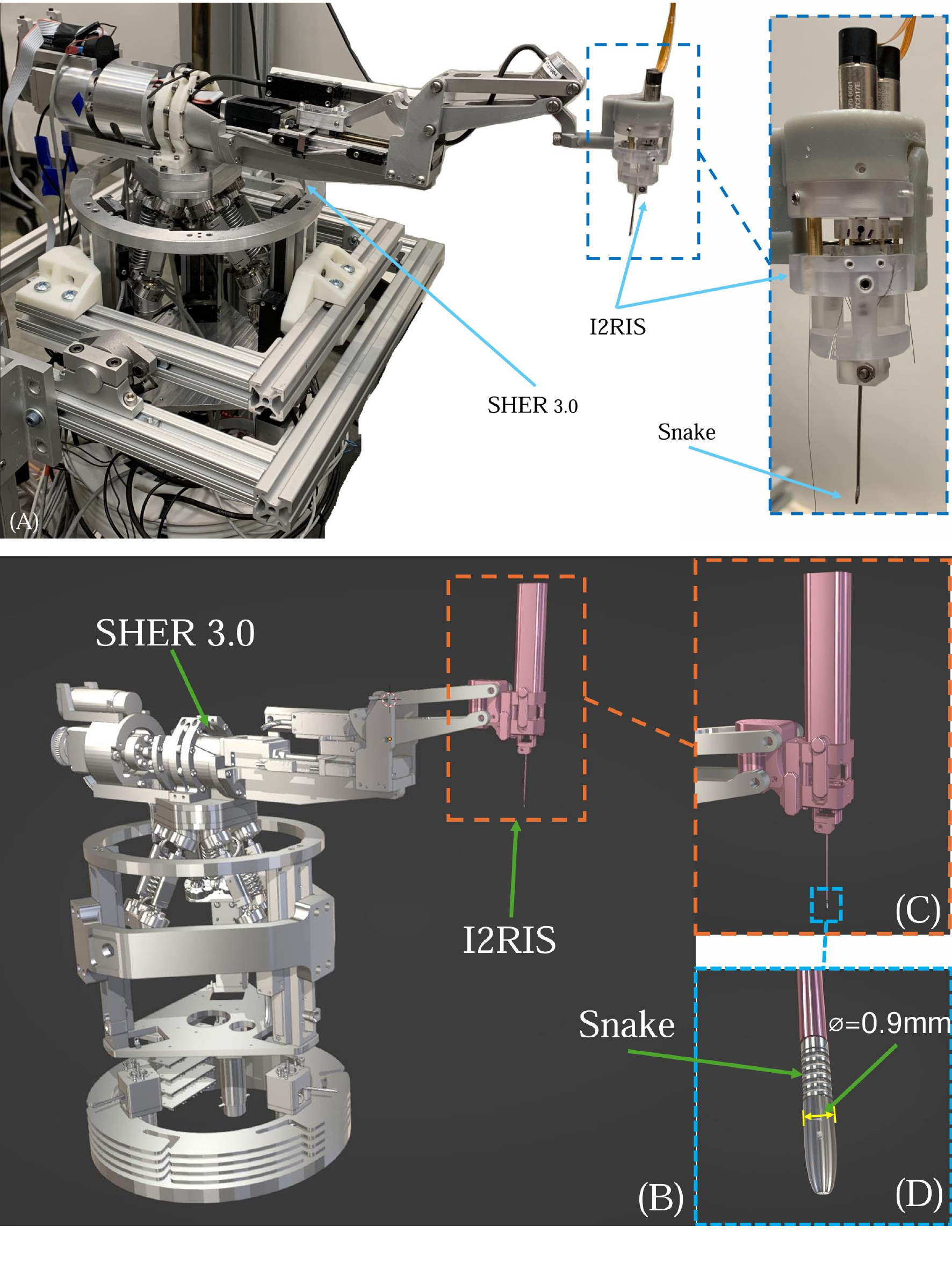}
      \caption{The Steady-Hand Eye Robot
(SHER) 3.0 integrated with the I$^2$RIS robot: (A) the real robots, where I$^2$RIS is attached to the end effector of SHER 3.0. (B) The digital twin developed in AMBF \cite{munawar2019asynchronous} provides a useful simulation environment for testing control strategies. (C) The digital twin of the I$^2$RIS alone, and (D) the articulated segment and the gripper of the I$^2$RIS.} 
      \label{fig:SHER_IRIS}
   \end{figure}

% \vfill

Several robotic systems were developed to minimize human error, including the Steady-Hand Eye Robot (SHER) 2.0 \cite{fleming2008cooperative}, SHER 2.1 \cite{zhao2023human}, the PRECEYES Surgical System \cite{meenink2013robot}, and the Robot-Assisted Minimally Invasive Eye Surgery (iRAM!S) \cite{nasseri2013introduction}. SHER 3.0 \cite{alamdar2023steady} (see Fig. \ref{fig:SHER_IRIS}), a next-generation cooperative control robotic system, provides enhanced tremor reduction, precise localization, easy integration into surgical environments, and reduced costs and physical footprint. Yet, higher dexterity at the distal end is advantageous in enabling ERM peeling at an optimum approach angle and accessing hard-to-reach structures in the eye.  

Continuum robots, known for their versatility, flexibility, and small size, showed great promise in Minimally Invasive Surgeries (MIS), reducing invasive operations, postoperative pain, surgery times, and costs \cite{ iqbal2025continuum}. These promising capabilities could benefit ophthalmic surgeries as well. For example, the Improved Integrated Robotic Intraocular Snake (I$^{2}$RIS) \cite{jinno2021improved} adds a small wrist-like mechanism with pitch and yaw degrees of freedom (DoFs) and gripping. Integrating I$^{2}$RIS with SHER 3.0 (see Fig. \ref{fig:SHER_IRIS}) can increase the overall dexterity of the SHER 3.0 during membrane peeling procedures. As a result, the combination of SHER 3.0 and I$^{2}$RIS can optimize surgical accuracy and efficiency while minimizing patient harm caused by human error. Moreover, the I$^{2}$RIS can provide fine and delicate manipulation and optimize the approach angle for tissue grasp, resulting in reduced tractional forces and hemorrhage during ERM peeling. However, developing kinematic and dynamic models for continuum robots remains challenging due to their complexity compared to traditional rigid-link models, as they involve a large number of DoFs \cite{russo2023continuum, rezaeian2023telescopic, ceja2024towards}. Furthermore, embedding sensors in continuum robots is challenging due to their small size and limited space \cite{da2020challenges}. Without reliable feedback from robot configuration, real-time estimation based on complex model-based algorithms becomes challenging, particularly in the presence of uncertainties \cite{russo2023continuum, amirkhani2023design}.

Classical model-based control methods, such as Cosserat rod theory and rigid link models, are applied to the kinematic modeling of continuum robots. However, these methods involve trade-off between accuracy and computational complexity, especially under environmental interaction or dynamic contact \cite{russo2023continuum, da2020challenges}. The piecewise constant curvature (PCC) model \cite{lopez2023review} requires ideal assumptions, such as neglecting friction and the mass of the robot, which reduces the accuracy of the estimates \cite{wang2021hybrid}.

Model-free control methods were developed to overcome these limitations. For example, methods such as local Gaussian Regression \cite{fang2022efficient}, Jacobian matrix updates \cite{yip2016model} and reinforcement learning \cite{kargin2023reinforcement, 10585785} provide adaptation to the system. Additionally, researchers employed adaptive algorithms during runtime to control manipulators in combination with Model Predictive Control (MPC). This approach aimed to improve the dynamic performance and reduce steady-state error in continuum robots under various conditions, such as restrictions on design and dynamics \cite{hyatt2020model}. For example, Nonlinear Model Predictive Control (NMPC) is used to control concentric tube robots (CTRs) \cite{khadem2020autonomous}. Pneumatically actuated continuum robots are controlled by Nonlinear Evolutionary Model Predictive Control (NEMPC), which incorporates Deep Neural Networks (DNN) \cite{hyatt2020real} for dynamics approximation. The main drawback of MPC is its computational demand, resulting in significantly longer processing times for each sampling period \cite{amouri2022nonlinear}.

Model Predictive Path Integral (MPPI) control is an advanced, model-based optimal control framework that utilizes stochastic trajectory sampling to generate fast, near-optimal control inputs for complex nonlinear and stochastic systems \cite{williams2017model, wang2019information, akshay2020hamiltonian}. Unlike traditional approaches like MPC, MPPI imposes no specific restrictions on the form of the state cost function \cite{pezzato2023sampling}, making it more flexible for a wide range of applications. By continuously generating a series of sampled trajectories through the system's dynamic model and evaluating their associated costs, MPPI computes an open-loop control sequence by integrating the costs along these paths. One key advantage of MPPI over MPC is its reliance on a gradient-free, sample-based numerical strategy, which makes it particularly effective for handling nonlinear dynamics and non-convex cost functions, where conventional optimization techniques like MPC may struggle. MPPI also benefits from its computational efficiency in certain scenarios, as it avoids solving complex optimization problems at each time step, instead relying on stochastic sampling, which can be parallelized and, using a graphics processing unit (GPU) \cite{williams2015gpu}, adapted to real-time applications. 

\begin{figure}[t!]
    \centering
    \includegraphics[width=0.49\textwidth]{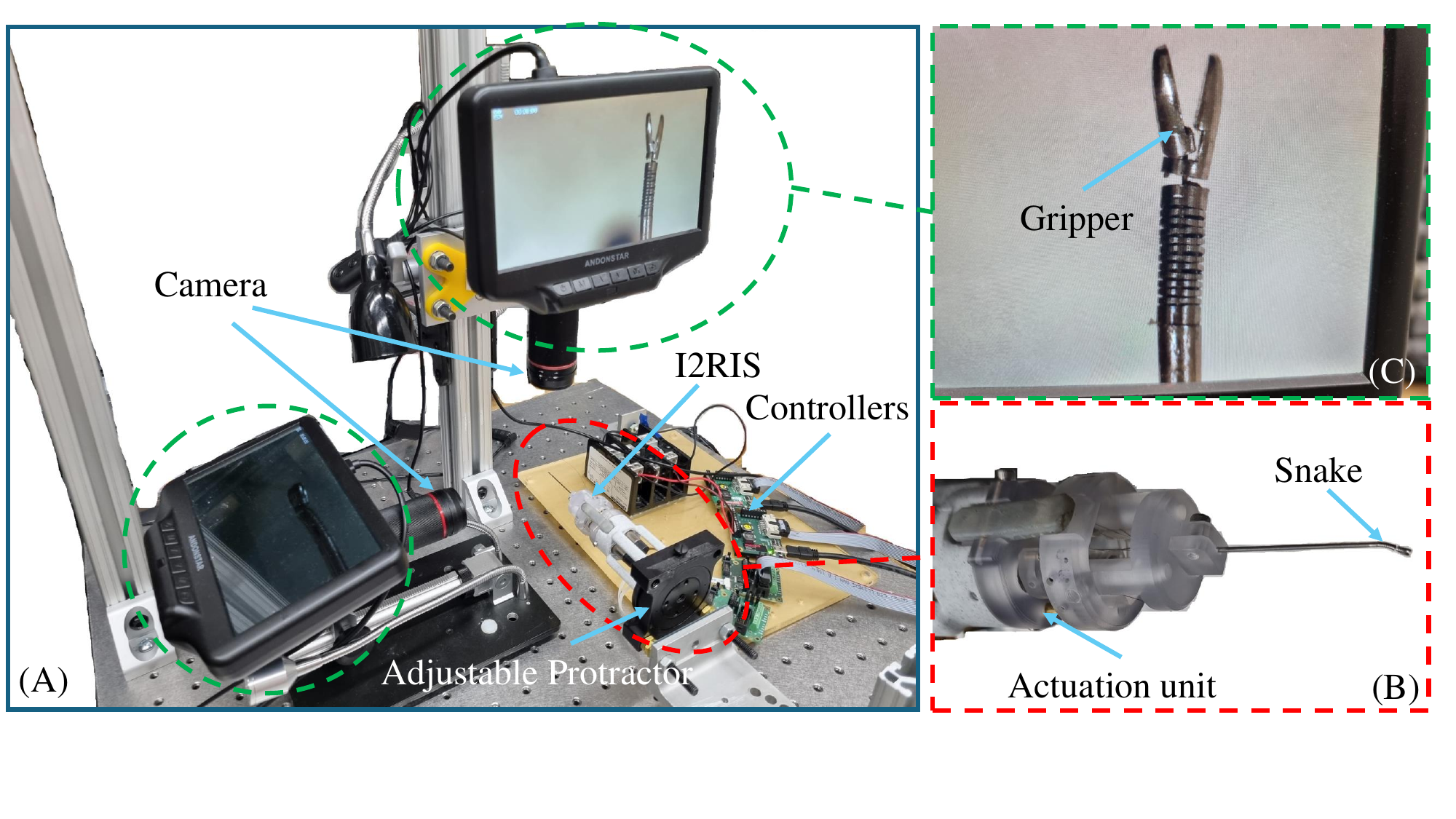}
      \caption{ (A) An overview of the experimental setup, (B) the I$^2$RIS actuation unit, and (C) an image of the articulated segment and the gripper of the I$^2$RIS captured by the camera.} 
      \label{fig:experiment_setup}
   \end{figure}

In this paper, we simulated an MPPI control algorithm on a data-driven probabilistic model of the I$^2$RIS based on Gaussian mixture models (GMM) and Gaussian mixture regression (GMR) \cite{esfandiari2024data}. The contributions of this work are as follows: 

\begin{itemize}
    \item We employed a radial basis function identifier to estimate the states of the I$^2$RIS, pitch, and yaw DoFs coming from the GMM-GMR model. 
    \item We designed an extended Kalman filter to update the weights of the RBF identifier to adapt to the unknown external disturbances and synthesized environmental loads added to the GMM-GMR model, trying to improve the adaption of the robot during real tissue interaction.  
    \item We presented a discrete-time MPPI controller for the I$^2$RIS model. The MPPI used the estimated states of the I$^2$RIS through the EKF-RBF identifier. We also compared the performance of the MPPI with an MPC in a trajectory tracking problem.   
\end{itemize}

 Of note, the previously developed GMM-GMR model is trained offline using experimental data from the robot's free motion, whose accuracy may be significantly impacted in the presence of unknown environmental loads during real tissue interaction. A Kalman filter can provide learning speeds faster than conventional RBF using gradient descent methods \cite{simon2002training}.

This paper is organized as follows. Section II describes the materials and methods. Section III presents and discusses the simulation results. Section IV concludes the paper.

%%%%%%%%%%%%%%%%%%%%%%%%% METHODS %%%%%%%%%%%%%%%%%%%%%%%%%%
\section{MATERIALS AND METHODS}

 To analyze the performance of the MPPI in controlling the I$^2$RIS robot, we used a probabilistic model of the robot developed in a previous study \cite{esfandiari2024data}. This model was based on Gaussian mixture models (GMM) and Gaussian mixture regression (GMR) with 15 Gaussian functions ($K=15$) whose parameters (prior probability $\pi_k$, mean value $\mu_k$, and covariance matrix $\Sigma_k$ for the $k$-th Gaussian function) were optimized by the Expectation Maximization (EM) algorithm \cite{moon1996expectation} using experimental data collected from the I$^2$RIS control inputs (motor angular position) and the output states (pitch and yaw DoFs) measured by two microscopes (see Fig. \ref{fig:experiment_setup}) \cite{esfandiari2024data} .

\subsection{System Model}

We assume that the robot motion could be characterized by a discrete-time nonlinear dynamic between the control action (motor angular position in $rad$) at time step $k$, denoted as $u_k \in \mathbb{R}^m$, and the snake tip bending angle $x_k \in \mathbb{R}^n$ for each of the pitch and yaw DoFs as follows:
\begin{equation}
    x_{k+1} = x_k + f(x_k, u_k) u_k + \mathcal{W}_k \\ 
   \label{eq:process_model}
\end{equation}
\begin{equation}
     z_k = H_k x_k + \mathcal{V}_k
     \label{eq:sensor_model}
\end{equation}
where \( f(x_k, u_k) \) is an unknown nonlinear function representing the system dynamics, $\mathcal{W}_k \sim \mathcal{N}(0, Q_k)$ is the process noise, a zero-mean Gaussian random sequence with covariance $Q_k = \mathbb{E}(\mathcal{W}_k \mathcal{W}_k^T)$, which represents the environmental disturbances and unmodeled dynamics, $z_k$ is an observation (or measurement) of the true state $x_k$ at time $k$, $H_k$ is the observation model, and $\mathcal{V}_k \sim \mathcal{N}(0, R_k)$ is the observation noise, a zero-mean Gaussian white noise with covariance $R_k = \mathbb{E}(\mathcal{V}_k \mathcal{V}_k^T)$ that is uncorrelated with the process noise, i.e. $\mathbb{E}(\mathcal{W}_j \mathcal{V}_k^T)=0$ ($\forall j,k$) \cite{stengel1994optimal}. Based on the current system setup with two monocular cameras capturing the pitch and yaw motions (see Fig. \ref{fig:experiment_setup}), we can assume that the states are fully observable ($H_k = I$) when the snake robot is outside the eye. If inside the eye, we may still have a full or partial observation of the tip orientation using optical coherence tomography (OCT) imaging \cite{ arikan2024real, arikan2024towards}.

We also assume that the pitch and yaw motions could be decoupled, each of them modeled by \eqref{eq:process_model} ($m=n=1$) with a different nonlinear function $f(x_k, u_k)$ depending on their cable tension and friction conditions.  

\subsection{RBF Identifier}

Our objective is to identify the nonlinear function $f(x_k, u_k)$ to estimate the robot states $x_k$, trying to control it toward a desired state $x_d$. To this end, we use a radial basis function due to its universal function approximation capability and fast learning process \cite{simon2002training}. The RBF approximation of the function \( f(x_k, u_k) \) could be defined as:   
\begin{equation}
\hat{f}(x_k, u_k, w_k) = \sum_{i=1}^{N} w_i \cdot \phi_i(x_k, u_k)
\label{eq:f_hat_RBF}
% \label{eq:f_hat_RBF}
\end{equation}
where \( \hat{f}(x_k, u_k) \) is the RBF estimation of \( f(x_k, u_k) \), \( N \) is the number of basis functions, \( w_i \) are the weights associated with each basis function, \( \phi_i(x, u) = \exp \left( -\frac{\| [x, u] - c_i \|^2}{2 \sigma_i^2} \right) \) is the \( i \)-th basis function with center \( c_i \in \mathbb{R}^{m+n} \) and width \( \sigma_i \), and $[x,u]$ is the state-input pair vector.  
The centers \( c_i \) are determined using \( k \)-means clustering algorithm, and the widths \( \sigma_i \) are set based on the maximum distance between the cluster centers \cite{yang2018fast}.

\subsection{Extended Kalman Filter for RBF Weights Update}

To adapt the robot identifier to the snake robot's external disturbances and environmental loads, the RBF weights \( w_i \) are updated online using an extended Kalman filter. The state of the Kalman filter is defined as the vector of RBF weights, ${w}_k = [w_1, w_2, \dots, w_N]^\top$.  
% \begin{equation}
% {w}_k = [w_1, w_2, \dots, w_N]^\top
% \end{equation}

The system model for the Kalman filter is the dynamics of the RBF in discrete time as follows:   
\begin{equation}
{w}_{k+1} = {w}_k + {\eta}_k
\label{eq:w_w_eta}
\end{equation}
\begin{equation}
z_k = (x_{k+1} - x_k)/u_k = f(x_k, u_k, w_k)= {\phi}_k^\top {w}_k + \nu_k
\label{eq:z_phi_nu}
\end{equation}
in which \eqref{eq:w_w_eta} specifies that the weights of the RBF are modeled as a stationary process with a process noise \( {\eta}_k \sim \mathcal{N}(0, Q'_k) \) with covariance matrix \( Q'_k \), and \eqref{eq:z_phi_nu} is the observation (measurement) model given by the predicted state difference in which $z_k$ is the desired response of the RBF identifier $\hat{f}_k(.)$ in \eqref{eq:f_hat_RBF}  \cite{haykin2004kalman, sum1999kalman,  pesce2020radial}. Here, \( {\phi}_k \in \mathbb{R}^N \) is the vector of RBF basis functions, and \( \nu_k \sim \mathcal{N}(0, R'_k) \) is the measurement noise with covariance \( R'_k \).

The Kalman filter update law for the states ${w}_k$ (RBF weights) based on the observed data is formulated as follows:
\begin{enumerate}
    \item Prediction Step: 
    \begin{align}
 \quad \hat{{w}}_{k+1|k} &= \hat{{w}}_k \\
P_{k+1|k} &= P_k + Q'_k \label{eq:P_P_Q_kalman}
    \end{align}
    \item Measurement Update:
    \begin{align}
 \quad {K}_k &= P_{k+1|k} H_k \left( H_k^\top P_{k+1|k} H_k + R'_k \right)^{-1} \label{eq:K_P_H_kalman} \\
\hat{{w}}_{k+1} &= \hat{{w}}_{k+1|k} + {K}_k \left( z_k - H_k^\top \hat{{w}}_{k+1|k} \right) \\
P_{k+1} &= \left( I - {K}_k H_k^\top \right) P_{k+1|k} \\ 
\mathcal{H}_k = & \frac{\partial \hat{f}(x_k,u_k, w_k)}{\partial {w_k}} = {\phi}_k 
\label{eq:H_k_phi}
\end{align}
\end{enumerate}
where \( \hat{{w}}_{k+1} \) is the updated estimate of the weights, \( P_{k+1} \) is the updated error covariance matrix, and \( {K}_k \) is the Kalman gain.

\subsection{MPPI Control}

The model predictive path integral is a model-based optimal control framework that uses stochastic sampling of trajectories to generate fast optimal control solutions for nonlinear and stochastic systems without restrictions on
the state cost function form. MPPI leverages a sampling-based control strategy that constantly generates a sequence of sampled trajectories using a dynamic system model and computes the costs of the sampled trajectories. An open-loop optimal control sequence is then generated by evaluating the path integral of the sampled trajectories \cite{williams2017model}. An MPPI control algorithm could be implemented on a dynamic system by taking the following steps:    

\subsubsection{Control Sampling}
At each time step, MPPI generates \( M \) sampled trajectories with a finite time horizon \( H \) by rolling out \( M \) random sequence of control inputs \( \{ u_k^i \}_{i=1}^{M} \sim \mathcal{N}(\mu, \Sigma) \), where $i=1,\cdots, M$ stands for the index of the $i$-th sampled trajectory. 
At each time step, the control sequence for the $i$-th sampled trajectory, \( \{ u_k^i \}_{k=0}^{H-1} \), is sampled over a finite horizon $k=0,\cdots, H-1$ from the following distribution \cite{tao2023rrt}:
\begin{equation}
u_k^i = u_k + \sigma_u \cdot \mathcal{N}(0, I)
\label{eq:u_rollout}
\end{equation}
where $\sigma_u$ is the variance of control sampling noise determining how wide the random samples of the control inputs are. The system state for the $i$-th trajectory, \( \{ x_k^i \}_{k=0}^{H-1} \), corresponding to each control sequence in \eqref{eq:u_rollout} is propagated using the system dynamics as follows:

\begin{equation}
x_{k+1}^i = x_k^i + \hat{f}(x_k^i, u_k^i) \cdot u_k^i
\end{equation}
where $\hat{f}(x_k^i, u_k^i)$ is estimated by the RBF identifier with EKF update \eqref{eq:f_hat_RBF}-\eqref{eq:H_k_phi} to adapt to the external disturbances and unmodeled dynamics.

\subsubsection{Cost Function}
The goal of the MPPI as a nonlinear model predictive control algorithm is to generate an optimal control policy $u_k$ that minimizes the following cost function over a finite horizon \( H \):
\begin{align}
\min_{u_k} \quad J =  & \Phi(x_H) +  \sum_{k=0}^{H-1} c(x_k, u_k) \\ 
s.t. \quad  x_{k+1} = & x_k + f(x_k, u_k)u_k + \mathcal{W}_k \nonumber
\end{align}
where $\Phi(x_H)$ is a quadratically bounded terminal cost and $c(x_k, u_k)$ is the stage cost defined as:
\begin{equation}
\Phi(x_H) = (x_H - x_{\text{desired}})^\top Q_H (x_H - x_{\text{desired}})
\end{equation}
\begin{equation}
c(x_k, u_k) = (x_k - x_{\text{desired}})^\top Q (x_k - x_{\text{desired}}) + u_k^\top R u_k
\end{equation}
in which \( Q \) is positive semidefinite, and \( Q_H \) and \( R \) are positive definite weighting matrices \cite{kohler2023stability}. To follow a desired trajectory, we select $T$ discrete points (states) along the trajectory, and for every iteration, we set the current state of the robot $x_k$ as $x_{\text{current}}$ or the initial state of the optimization problem, and the next point on the trajectory $x_{\text{traj}}$ as the $x_{\text{desired}}$. We keep updating $x_{\text{current}}$ and $x_{\text{desired}}$ until all $T$ points on the trajectory are tracked.

\begin{algorithm}[t]
\caption{MPPI with RBF Identifier and EKF Update}
\label{alg:mppi_rbfnn_kalman}
\KwIn{$x_0, u_0, x_{\text{desired}}, H, M, \lambda, \sigma_u, c_i, \sigma_i, w_i, P_0, Q'_0, R'_0$}
\KwOut{Optimal control sequence $u_k$}
Initialize $x_{\text{current}} \gets x_0$, $x_{\text{desired}} \gets x_{\text{traj}}$ ; (Set the next state on the trajectory
as the desired state)\\
Discretize trajectory into $T$ points\;
\While{Not all $T$ points are tracked}{
    Sample $M$ control sequences: $u_k^i = u_{k-1}^i + \sigma_u \mathcal{N}(0,1)$\;
    \For{$i \gets 1$ \KwTo $M$}{
        Initialize $x_{\text{temp}}^i \gets x_{\text{current}}$, $J^i \gets 0$\;
        \For{$k \gets 1$ \KwTo $H-1$}{
            Compute RBF basis $\phi_j(x_{\text{temp}}^i, u_k^i)$ for $j=1,\dots,N$\;
            Estimate dynamics $\hat{f}(x^i_{\text{temp}}, u^i_k) = \sum_{j=1}^N w_j \phi_j(x_{\text{temp}}^i, u_k^i)$\;
            Update state $x_{\text{next}}^i = x_{\text{temp}}^i + \hat{f}(x_{\text{temp}}^i, u_k^i) \cdot u_k^i$\;
            Compute stage cost $J^i \gets J^i + (x_{\text{next}}^i - x_{\text{desired}})^\top Q (x_{\text{next}}^i - x_{\text{desired}}) + u_k^{i\top} R u_k^i$\;
            Update $x_{\text{temp}}^i \gets x_{\text{next}}^i$\;
        }
        Compute terminal cost \\ $J^i \gets J^i + (x_{\text{next}}^i - x_{\text{desired}})^\top Q_H (x_{\text{next}}^i - x_{\text{desired}})$\;
        Compute control weights $\rho^i = \exp\left( -\frac{J^i}{\lambda} \right)$\;
    }
    Compute optimal control $u_k = \frac{\sum_{i=1}^M \rho^i u_k^i}{\sum_{i=1}^M \rho^i}$\;
    Apply $u_k$ to the robot model and measure $x_{k+1}$\;
    \textbf{Kalman Update:}\\
    Prediction error $e_k = \frac{x_{k+1} - x_k}{u_k} - \phi_k^\top \hat{w}_{k+1|k}$\;
    Kalman gain $K_k = P_{k+1|k} \phi_k \left( \phi_k^\top P_{k+1|k} \phi_k + R'_k \right)^{-1}$\;
    Update weights $\hat{w}_{k+1} = \hat{w}_{k+1|k} + K_k e_k$\;
    Update covariance $P_{k+1} = (I - K_k \phi_k^\top) P_{k+1|k}$\;
    Update states $x_{\text{current}} \gets x_{k+1}$, $x_{\text{desired}} \gets x_{\text{traj}}$\;
}
\end{algorithm}

% \vspace{-15pt}

\subsubsection{Cost Evaluation}
MPPI solves the above optimal control problem numerically based on a sample-based gradient-free stochastic strategy, making it suitable for systems with nonlinear dynamics and non-convex cost functions. For each sampled trajectory, the cumulative cost is computed as:
\begin{align}
 J^i = & \Phi(x^i_H) + \sum_{k=0}^{H-1} c(x_k^i, u_k^i) \\ 
 s.t. \quad  x_{k+1}^i = & x_k^i + \hat{f}(x_k^i, u_k^i, {w}_k)u_k^i + \nu_k \nonumber \\ 
    u_k^i = &  u_k + \sigma_u \cdot \mathcal{N}(0, I) \nonumber 
\end{align}

To ensure that those sampled control sequences $u_k^i$ that resulted in sampled trajectories with higher costs $J^i$ should have less importance in determining the optimal control solution $u_k$, we define control weights $\rho^i$ for the $i$-th trajectory as the exponential of the negative cost on that trajectory, computed as follows:
\begin{equation}
\rho^i = \exp \left( -\frac{J^i}{\lambda} \right)
\label{eq:w_MPPI}
\end{equation}
where \( \lambda \) is a temperature parameter.

\subsubsection{Control Update}
The optimal control solution of the MPPI algorithm is updated by averaging the sampled control sequences weighted by the exponentiated cost in \eqref{eq:w_MPPI} as follows:
\begin{equation}
u_k = \frac{\sum_{i=1}^{M} \rho^i u_k^i}{\sum_{i=1}^{M} \rho^i}
\end{equation}

This optimal control sequence is then applied to the system. For a trajectory tracking problem, Algorithm \ref{alg:mppi_rbfnn_kalman} summarizes the proposed MPPI controller and the RBF identifier with EKF weight updates.

\section{RESULTS AND DISCUSSIONS}

\begin{figure}[t!]
	\centering
    \subfloat[]{
    \centering
    \includegraphics[width=0.43\textwidth]{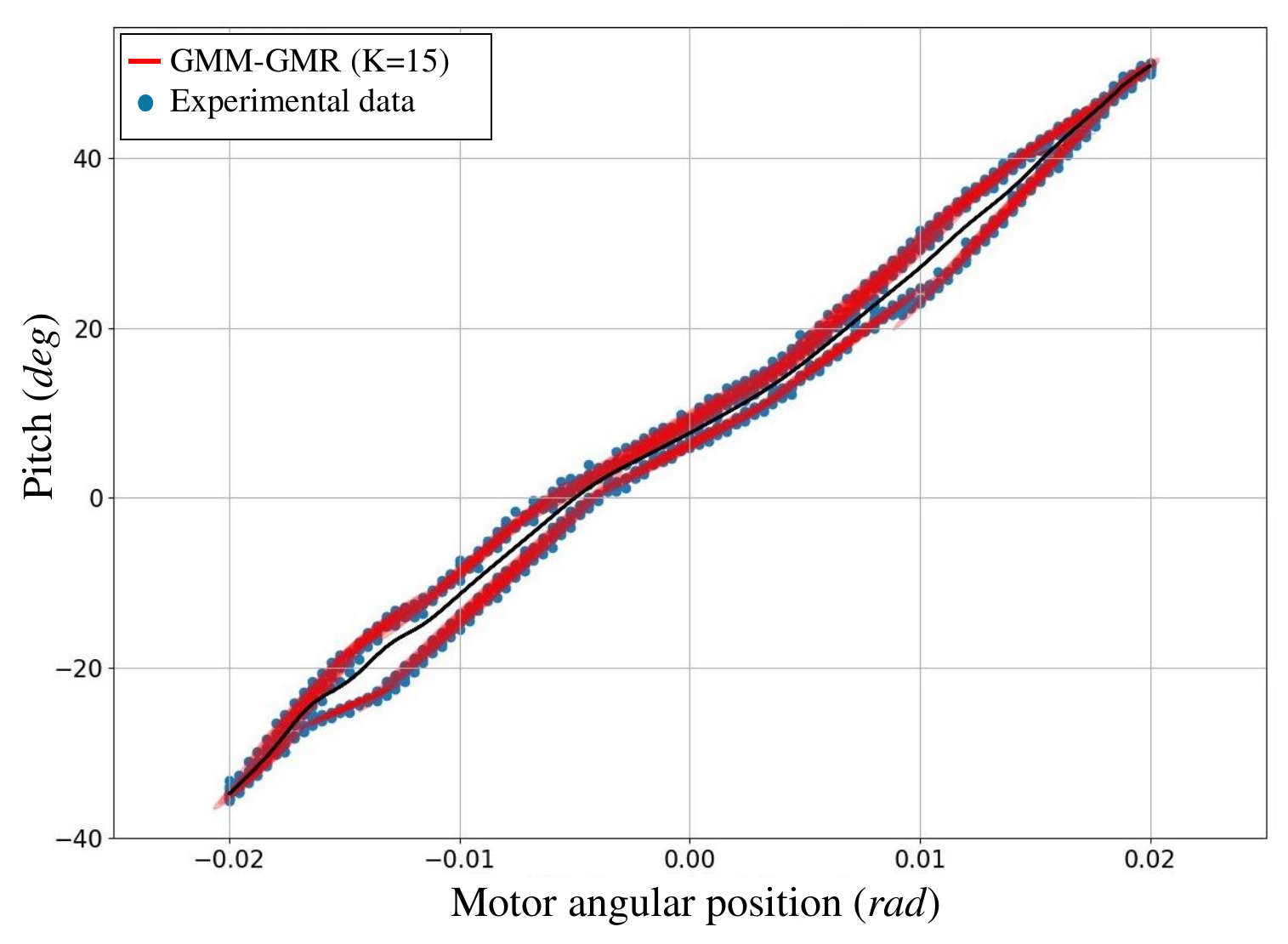}
    \label{fig:SideViewPitch_GMR}}  \\ 
    \centering
    \subfloat[]{
    \centering
    \includegraphics[width=0.43\textwidth]{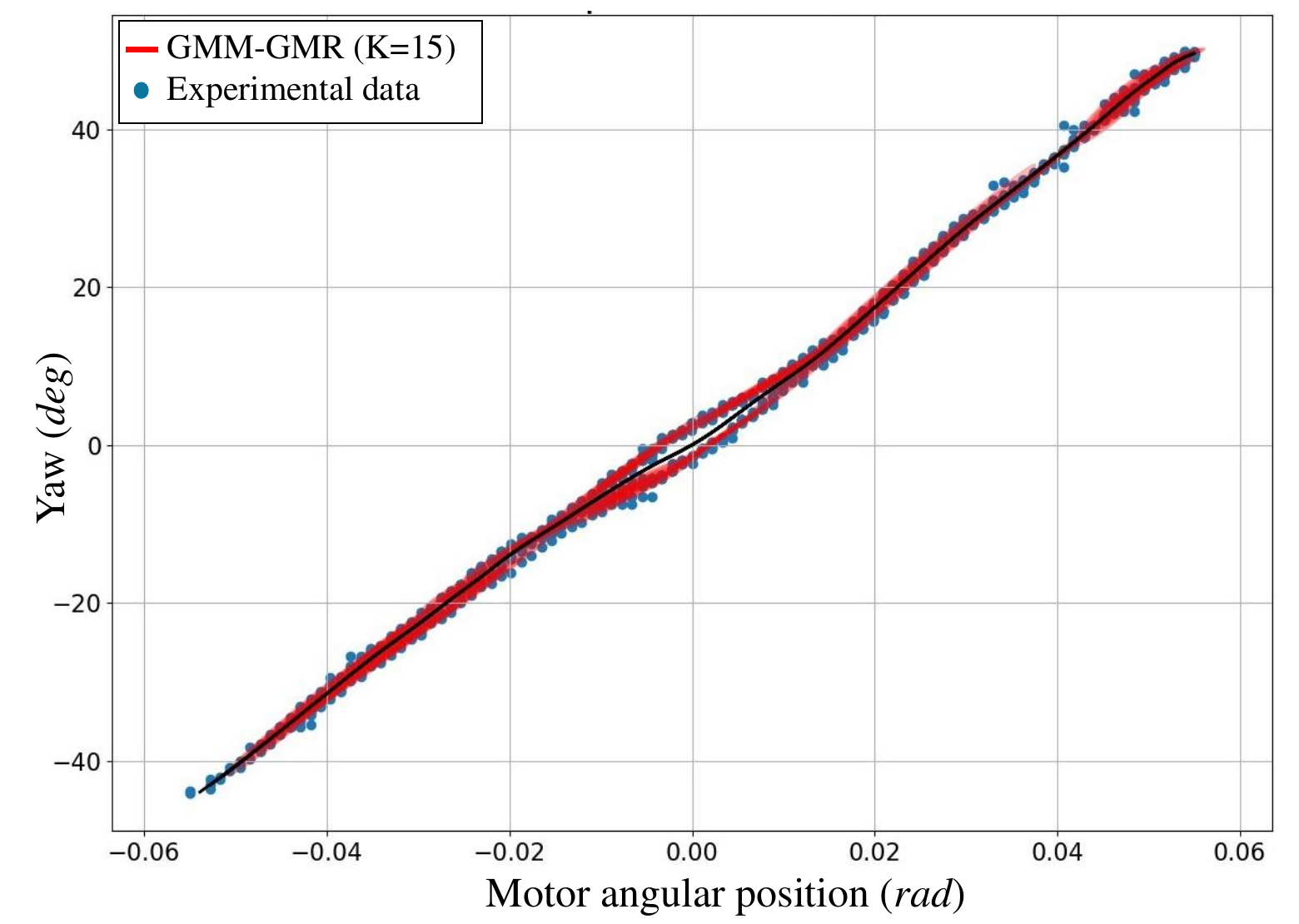}
    \label{fig:TopViewYaw_GMR}}
	\caption{Results of the GMM-GMR model in estimating the pitch (a) and yaw (b) degrees for different values of motor shaft angles (rad). A total number of $K=15$ Gaussian functions are used in this model. The blue dots are experimental data points for the input-output pairs $(u_k, x_k)$, and the red curve is the GMM-GMR model output.}
\label{fig:GMM_GMR_Model}
\end{figure}

In this study, we evaluate the performance of the MPPI controller on a probabilistic (GMM-GMR) model of the I$^2$RIS robot developed in a previous study \cite{esfandiari2024data} (see Fig. \ref{fig:GMM_GMR_Model}). We identify the output of this GMM-GMR model by an RBF whose weights are updated online by an extended Kalman filter to adapt to the potential external disturbances and environmental loads caused during robot-tissue interactions. These external disturbances are modeled as $\mathcal{W}_k \sim \mathcal{N}(0, Q_k)$ with a variable noise covariance $Q_k = 0.02 \, \norm{x_k}$ and the unknown environmental loads are modeled as $0.03 \,  x_k \, \sin(10 x_k)$, where $x_k$ is the robot state coming from the GMM-GMR model. We intentionally made these disturbances dependent on the robot state $x_k$, as it is seen in the real snake robot that the more it is bent, the more nonlinear behavior is observed due to the increase in the cables' tension and friction \cite{esfandiari2024data}.

The RBF identifier has 10 basis functions ($N=10$), and the weights are initialized by a multivariate normal distribution with mean $\textbf{0}_{10\times 1}$ and covariance $\textbf{I}_{10 \times 10}$ as $w_0 \sim \mathcal{N}(\textbf{0},\textbf{I})$.

Due to the Kalman filter's computational superiority over gradient descent methods \cite{simon2002training}, we used an extended Kalman filter to estimate and update the RBF weights. The extended Kalman filter gains are initialized with $P_0= I_{N\times N}$ for the initial estimation of the error covariance, $Q'_0 = 0.07 \, I_{N \times N}$ for the process noise covariance, and $R'_0 = 0.1$ for the measurement noise covariance with $N=10$. 

% Fig. \ref{fig:Kalman_RBFNN_estimation} shows an example of the EKF-RBF estimation of the pitch degree from the GMM-GMR model of the snake with added noise for external disturbances and environmental loads.    

The parameters of the MPPI control algorithm used in this study are as follows. We used $M=20$ sampled trajectories over a horizon $H=20$, a temperature value of $\lambda= 0.01$, the variance of control noise $\sigma_u =0.005$, terminal cost weight $Q_H = 17$, and stage cost weights $Q=0.2$ and $R=0.6$.

 We also compared the MPPI controller with a conventional MPC \cite{esfandiari2019nonlinear} in the presence of external disturbances, measurement noise, and simulated environmental loads.

 % \begin{figure}[t!]
 %    \centering
 %    \includegraphics[width=0.51\textwidth]{figures/Kalman_RBFNN_estimation.png}
 %      \caption{ An example of the estimation of the robot model states (pitch DoF) by the RBF identifier with EKF update for RBF weights in the presence of external disturbances.   } 
 %      \label{fig:Kalman_RBFNN_estimation}
 %   \end{figure}

\begin{figure}[t!]
\centering
        \includegraphics[width= 0.47 \textwidth]{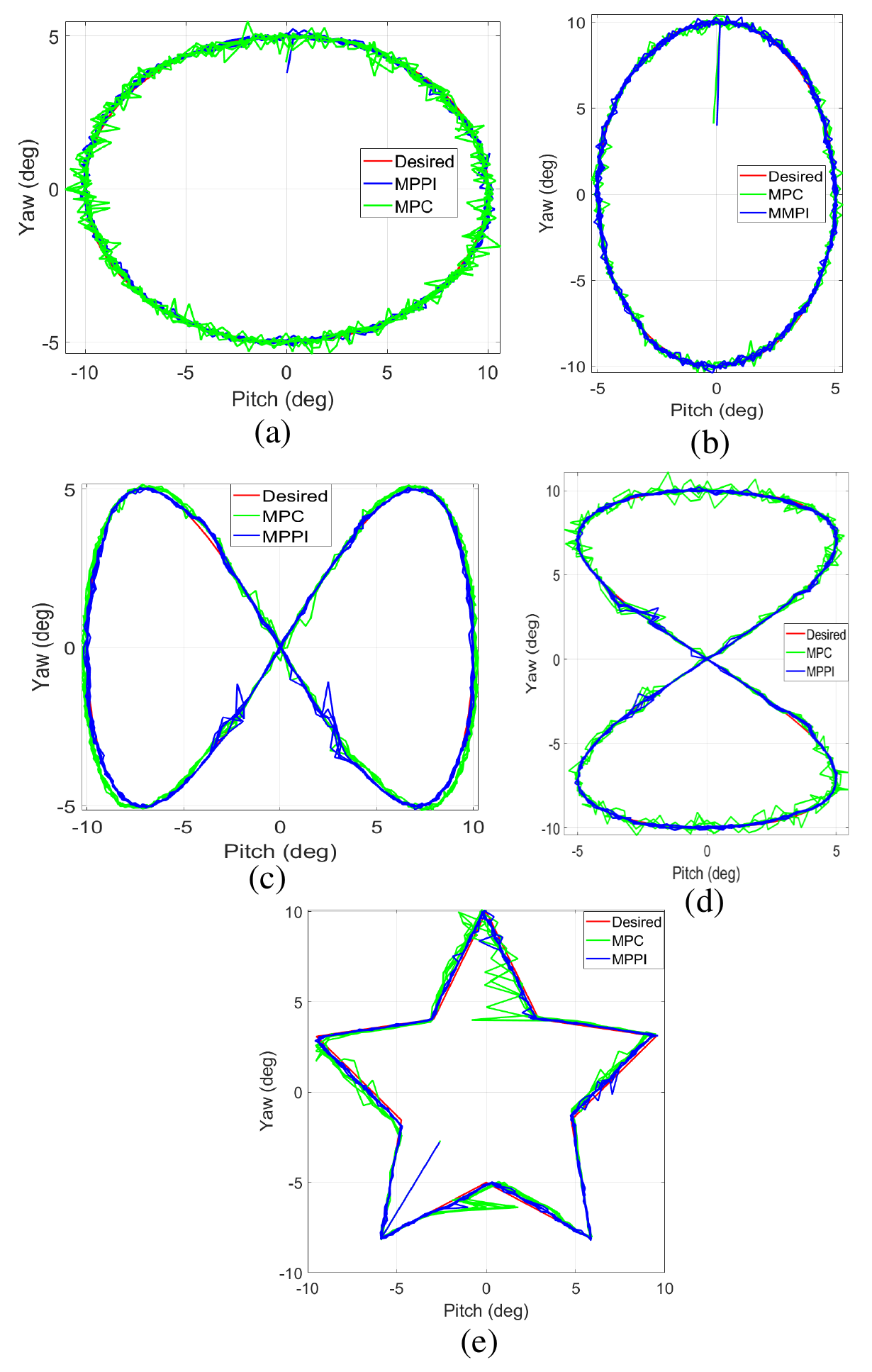}
    \caption{Trajectory tracking performance of the MPPI and MPC on the snake robot tip orientation model (GMM-GMR) in the presence of synthesized unknown external disturbances and environmental loads. The MPPI controller uses the estimation of the robot states from the RBF identifier, whose weights are updated using an extended Kalman. Five desired trajectories are defined, including (a) a horizontal oval with 10 and 5 deg for pitch and yaw, (b) a vertical oval, (c) a horizontal infinity sign, (d) a vertical infinity sign with a similar radius for pitch and yaw, and (e) a star with an inner and outer radius of 5 and 10 deg. Each desired trajectory (red) is tracked five times by both MPC (green) and MPPI (blue) controllers. MPPI showed a more robust behavior in the presence of unknown external disturbances with significant superiority in computational efficiency and speed.  }
     \label{fig:all_figs_MPC_MPPI_small}
\end{figure}

We defined five desired trajectories, horizontal and vertical ovals (\tikz \draw (0,0) ellipse (4pt and 2pt);), and infinity signs ($\infty$) with 5 and 10 $deg$ amplitudes for pitch and yaw DOFs, and a star ($\smallstar$) with an inner and outer radius of 5 and 10 $deg$. We discretized each trajectory into $T=200$ points and solved the optimal control problem between every two consecutive points, considering the $x_{\text{current}}$ one as the initial state and the next point to be reached on the trajectory, $x_{\text{traj}}$, as $x_{\text{desired}}$. Once the optimal control solution $u_k$ is found, it is applied to the robot's model (GMM-GMR) to move it toward the next desired point. The new state of the robot is then set as the current state for the next iteration, and another untracked point is set as a new desired state, and so on. This procedure is repeated until all $T$ points along a desired trajectory are tracked (see Algorithm \ref{alg:mppi_rbfnn_kalman}). The forceps' approach angle significantly affects successful ERM peeling \cite{chang2022efficiency}; hence, such maneuverabilities could provide an optimum approach angle and reduce potential hemorrhages. Fig. \ref{fig:all_figs_MPC_MPPI_small} shows the performance of the MPPI and MPC in tracking the desired trajectories. Each trajectory is tracked 5 times by each controller. The results are summarized in Table \ref{table:error_table_MPC_MPPI}.  

The simulations are performed on a CPU computer (Intel(R) Core(TM) i5-8250U 1.6GHz 1.8 GHz) in MATLAB (2021b), and the computational time for MPC and MPPI controllers are compared in Table \ref{table:error_table_MPC_MPPI}. It is observed that the MPPI is significantly faster than the MPC ($p<0.05$) and outperforms the MPC in terms of tracking RMSE. In a future study, we aim to implement this algorithm on both the digital twin and the real robot by further improving the computational speed of the MPPI using GPUs. Due to recent advances in parallel programming with GPUs, MPPI can generate optimal control solutions as fast as 50 Hz \cite{pravitra2020ℒ}, which is fast enough for our application.

% \begin{figure}[t!]
%     \centering
%     \includegraphics[width=0.51\textwidth]{figures/circle__traj_MPPI.png}
%       \caption{ } 
%       \label{fig:circle__traj_MPPI}
%    \end{figure}

\begin{table}[t!]
\centering
    \caption{Comparison of MPC and MPPI in tracking five trajectories in terms of computation time and RMSE (deg) for pitch and yaw.  }
    \includegraphics[width= 0.98 \columnwidth]{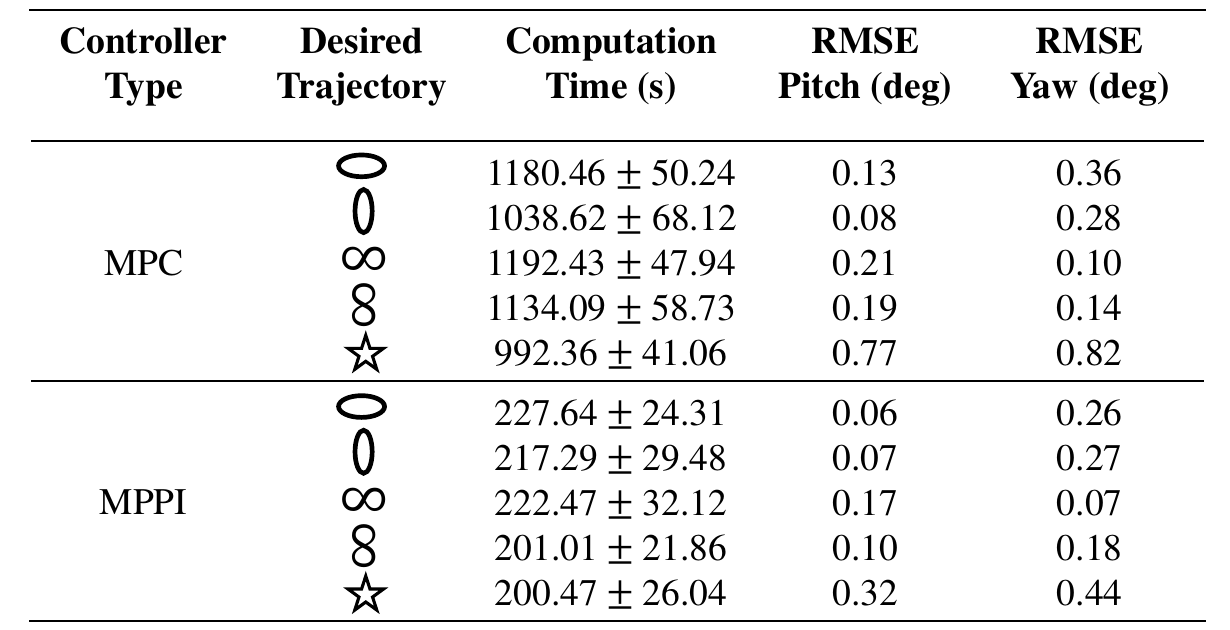}
\label{table:error_table_MPC_MPPI}
\end{table}

%%%%%%%%%%%%%%%%%%%%%%% CONCLUSIONS %%%%%%%%%%%%%%%%%%%%%%%%
\section{CONCLUSIONS}

In this study, we designed and simulated a model predictive path integral control algorithm for the I$^2$RIS robot. 
We used Gaussian mixture models and Gaussian mixture regression as a model for simulating the snake robot motion. The parameters of this GMM-GMR model were optimized using experimental data from the snake robot in a previous study. Here, we used a radial basis function for online identification of the robot states (the GMM-GMR model). The weights of the RBF are updated using an extended Kalman filter. Moreover, we compared the performance of the proposed MPPI controller with a conventional MPC in tracking five different trajectories.      
It is observed that the MPPI controller, along with the EKF-RBF identifier, showed more accurate and robust behavior compared to MPC in the presence of unknown external disturbances and environmental loads. Also, MPPI outperformed MPC in terms of computational time and returned optimal control solutions about 5 times faster than MPC. 
In the future, we plan to implement this control strategy on the I$^2$RIS robot using a GPU to take advantage of the parallel nature of MPPI for further improvement of computational performance.


%%%%%%%%%%%%%%%%%%%%%%%%%%%%%%%%%%%%%%%%%%%%%%%%%%%%%%%%%%%%%%%%%%%%%%%%%%%%%%%%

%%%%%%%%%%%%%%%%%%%%%%%%%%%%%%%%%%%%%%%%%%%%%%%%%%%%%%%%%%%%%%%%%%%%%%%%%%%%%%%%



% \section*{ACKNOWLEDGMENT}

%%%%%%%%%%%%%%%%%%%%%%%%%%%%%%%%%%%%%%%%%%%%%%%%%%%%%%%%%%%%%%%%%%%%%%%%%%%%%%%%

% \balance
 % \vfill

\bibliographystyle{IEEEtran}

\bibliography{bibliography}

\end{document}